\documentclass{article}

\usepackage[square,sort,comma,numbers]{natbib}
\usepackage[final]{nips_2018}
\usepackage{graphicx}  
\usepackage[utf8]{inputenc} 
\usepackage[T1]{fontenc}    
\usepackage{hyperref}       
\usepackage{amsmath}
\usepackage{color, colortbl}
\usepackage{url}            
\usepackage{booktabs}       
\usepackage{amsfonts}       
\usepackage{nicefrac}       
\usepackage{microtype}      
\usepackage{adjustbox}
\usepackage{wasysym}
\usepackage{textcomp}

\title{Unsupervised Pseudo-Labeling for Extractive Summarization on Electronic Health Records}

\author{Xiangan Liu\thanks{Equal Contribution}\quad\, Keyang Xu$^{*}$\quad Pengtao Xie\quad Eric Xing\\
Petuum Inc., Pittsburgh, USA\\
\texttt{\{xiangan.liu, keyang.xu, pengtao.xie, eric.xing\}@petuum.com }
}

\begin{document}

\maketitle

\begin{abstract}
Extractive summarization is very useful for physicians to better manage and digest Electronic Health Records (EHRs). However, the training of a supervised model requires disease-specific medical background and is thus very expensive. We studied how to utilize the intrinsic correlation between multiple EHRs to generate pseudo-labels and train a supervised model with no external annotation. Experiments on real-patient data validate that our model is effective in summarizing crucial disease-specific information for patients.
\end{abstract}

\section{Introduction}

Electronic Health Records (EHRs) are time-sensitive and patient-centered documents that make medical-related information available instantly and securely to authorized physicians. EHRs, however, are usually long and verbose. Physicians spend extensive time reading through contents written in unstructured or semi-structured natural languages to filter out irrelevant information and dig out the disease-specific problem lists such as past medical history, symptoms, and prescriptions. This process becomes even more time-consuming when one patient has several such records over many years. 

Previous studies~\cite{DBLP:journals/jamia/FordCSSC16, DBLP:journals/ijmi/LeePC18, liu2018learning} focused on how to better utilize and digest information from EHRs to enhance the efficiency of healthcare services. Summarization is one of those techniques that can be applied, which generally has two approaches: abstractive and extractive summarization. As abstractive summarization sometimes fails to capture factual details accurately as needed in medical settings, we consider extractive summarization to be more suitable. This method directly extracts a subset of data written by the medical experts as the summary. Unsupervised extractive summarization was first explored~\cite{brandow1995automatic, DBLP:conf/aaai/HeCBWZCH12, li2013using}. Due to the recent success of neural networks, supervised approaches become more popular for extractive summarization~\cite{ArumaeL18,cao2016attsum, DBLP:conf/acl/0001L16, kaageback2014extractive,DBLP:conf/aaai/NallapatiZZ17}. One obstacle for training a supervised model for extractive summarization in the medical domain, however, is the lack of labeled data, since annotations for EHRs require disease-specific medical background and can be very expensive. 

In this work, we tried to train a supervised model, which has a better generalization ability than unsupervised models, with no direct human annotations. We studied how to utilize the intrinsic correlation between multiple notes for a single patient to generate pseudo-labels and guide summarization and explore to answer the following three Research Questions (RQ):


\begin{itemize}
    \item \textbf{RQ1:} How to measure the quality of disease-specific summarization for the same patient?
    \item \textbf{RQ2:} Based on the criterion in RQ1, how can we generate pseudo-labels that best satisfy this criterion. 
    \item \textbf{RQ3:} Given pseudo-labels in RQ2, what model architecture should be used for summarization in a medical setting? 
\end{itemize}

\section{Our Approach}
\begin{figure*}[t]
\centering
\includegraphics[width=0.90\textwidth]{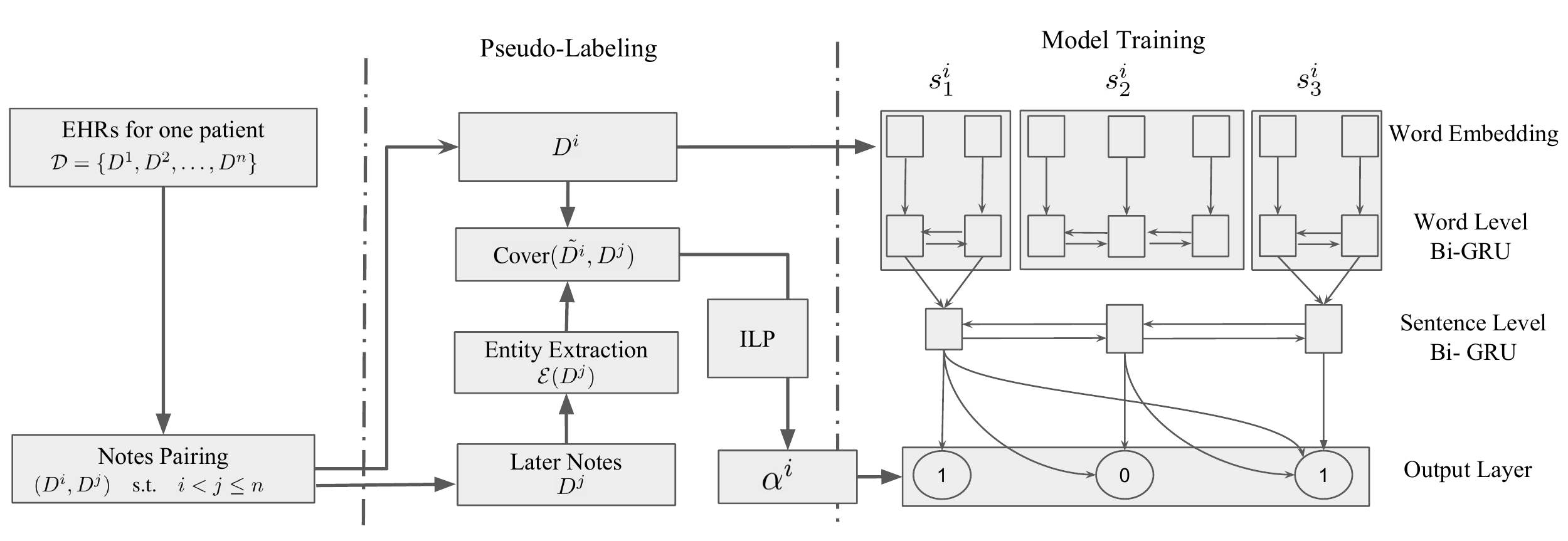}
\caption{The workflow of our unsupervised summarization method for EHRs. Left part is the notes pairing process. Middle part generates pseudo-labels, which are fed to train the neural model on right part.}
\label{fig:workflow}
\vspace{-1em}
\end{figure*}

\label{sec:method}
\subsection{Problem Definition}
Formally, we denote all EHRs recorded over time for one patient as $\mathcal{D}=\{D^1, D^2, \ldots, D^n\}$, where $D^1$ indicates the oldest note and $D^n$ is the newest one. For any note $D^i$, $1\leq i \leq n$, it contains a sequence of sentences as $D^i = \{s_1^i, s_2^i, \ldots, s_{|D^i|}^i\}$. Our task is to find a subset ${\tilde{D^i}} \subseteq D^i$, that best summarizes patient's information for a specific disease and also follows some length restrictions.  

For \textbf{RQ1}, we observed that when physicians read and summarize clinical notes, they focus more on medical entities that are related to a specific disease. For example, procedure entities such as "coronary artery bypass grafting" and "valve replacement" are very crucial for congestive heart failure; Lab test entities "hemoglobin" and "hematocrit" are informative for diagnosing anemia. We proposed to summarize clinical notes to cover more relevant entities. 

The entity set of $D^{i}$ is $\mathcal{E}(D^{i})$. However, hundreds of entities could exist in $\mathcal{E}(D^{i})$ and how to capture relevant ones remains to be a problem. Since this requires expensive medical expertise, we need a more efficient way to obtain patterns from clinical notes directly. We found that the important information in an early health record is usually  mentioned briefly again in later records, reminding physicians to pay attention for future treatments. This information includes but is not limited to lab test results, diagnoses and medication usages. Inspired by this, we defined a coverage score similar to Yu et al.'s study on TV series recap~\cite{DBLP:conf/emnlp/YuZM16}, to evaluate the quality of $\tilde{D^i}$ based on one of its later records $D^j$, where $i < j \leq n$, defined as, 
\begin{align}
    \text{Cover}(\tilde{D^i}, D^{j}) = \sum_{e \in \mathcal{E}(D^{j})} { \lambda(e)\max_{s_t^i \in D^{i}} \alpha_t^i \cdot \text{sim}(e, s_t^i)} 
\end{align}
where $e$ is one element of $\mathcal{E}{(D^j)}$ and $\lambda(e)$ is the importance measured by Inverse Document Frequency (IDF) score in the entire corpus. $\alpha^i$ is a binary vector indicating whether one sentence is selected for $\tilde{D_i}$. 
$\text{sim}(e,s_t^i)$ calculates the semantic similarity between $e$ and sentence $s_t^i$, as,
\begin{align}
    \text{sim}(e,s_t^i) =   \max_{{w} \in s_t^i}\, \text{cos}(\bar{e},\bar{w})
\end{align}
where $\bar{e}$ and $\bar{w}$ indicate the vectors that represent entity $e$ and word $w$ using average pooling of pre-trained word embeddings, which were trained on PubMed \footnote{https://www.ncbi.nlm.nih.gov/pubmed/} using skipgram \cite{DBLP:conf/nips/2013}. PubMed has a vocabulary that is more similar to EHR than general corpora and we used the abstracts of over 550,000 biomedical papers.  

\subsection{Pseudo-labeling with Integer Linear Programming}
Based on the definition introduced above, our target is to generate the binary label vector $\alpha^i$ for note $D_i$ using its later notes $D_j$, where $i < j \leq n$, which answers \textbf{RQ2}. In order to find the optimum $\alpha^i$ to maximize $\text{Cov}(\tilde{D^i}, D^{j})$, we used the Integer Linear Programming (ILP) for this optimization problem, as shown in Figure \ref{fig:workflow}. Ther target function for a pair of notes $D^i$ and $D^{j}$ as follows,
\begin{align}
    \max &  \sum _{e \in \mathcal{E}(D^{j})} \lambda(e) \max_{s_t \in D^i} \alpha^i_t \cdot \text{sim}(e,s_t^i) \quad  \text{s.t.} \quad \sum  \alpha_t^i \cdot |s_t^i| \leq L 
\end{align} 
where  $|s_j^i|$ is the number of words in $s_j^i$ and $L$ is a hyperparameter for length restriction. 
One notable problem is that $\max \alpha^i_t \cdot \text{sim}(c,s^i_t)$ is unsmooth, which requires unaffordable computational resources in real practice. To improve this, we used the log-sum-exp trick, which is a frequently adopted smooth approximation of the max function.
Also, we added two more length constraints $L_1$ and $L_2$ to make the optimization faster. The final optimization problem is defined as,
\begin{align}
    \max &  \sum _{e \in \mathcal{E}(D^{j})} \lambda(e) \cdot \text{log} \Big( \sum^{|D^i|}_{t=1}\text{exp}\big(\alpha^i_t \cdot \text{sim}(e,s_t^i)\big) \Big) \\
    \text{s.t.} \ & \sum^{|D^i|}_{t=1}  \alpha_t^i \cdot |s_t^i| \leq L, \quad  \text{and} \quad L_1 \leq \sum^{|D^i|}_{t=1}  \alpha_t^i \leq L_2, \quad \forall i 
\end{align}

\subsection{Summarization Model}
For \textbf{RQ3}, after we constructed training data with pseudo-labels $\alpha$, a supervised neural model was trained to summarize medical records. The model predicts the probability of each sentence being picked for summarization. The model consists of a two-layer bi-directional GRU~\cite{chung2014empirical}. As shown in the right part of Figure \ref{fig:workflow}, its first level Bi-GRU is on word level and generates sentence embedding. Taking it as input, the second layer Bi-GRU is on sentence level and final representation for each sentence is $h_t = [h_t^f, h_t^b]$. 

For the output layer for $t$-th sentence, we used a logistic function which contains several features, including \texttt{content}, \texttt{salience}, \texttt{novelty}, \texttt{position}. The \texttt{salience} reflects how representative current sentence is according to the entire note. The \texttt{novelty} helps us avoid redundancy. 
We can predict the probability of selecting the current sentence as:
\begin{align}
    P(y_t = 1| h_t,s_t,d) = \sigma( \underbrace{W_{c} h_{t}}_{\text{content}} +\underbrace{h_t^TW_sd}_{\text{salience}}  
    - \underbrace{h_t^TW_r\tanh(s_t)}_{\text{novelty}} 
    + \underbrace{W_{p} p_t}_{\text{position}}
    + b_1)  & 
\end{align}
where $p_t$ is the position embedding of current sentence's index $t$. $d$ is the representation of the entire note as $d = \tanh(\sum^{N_d}_{t=1}W_d h_t / N_d +b_2)$, where $N_d$ is the number of sentences in this note. The novelty feature $s_t$ is the weighted sum of the representations of already selected sentences, defined as $ s_t = \sum^{t-1}_{k=1}h_k\cdot P(y_k = 1| h_k,s_k,d)$.

For $\mathcal{D}$, cross-entropy loss was adopted to optimize the neural model over $W$ and $b$, as follows, 
\begin{align}
    \ell(W,b) & = -\sum^{|\mathcal{D}|}_{i=1}\sum^{|D^i|}_{t=1}\left(y_t^i\cdot \log P(y_t^i=1|h_t^i,s_t^i,d^i) 
     +(1-y_t^i) \cdot \log(1-P(y_t^i=1| h_t^i,s_t^i,d^i)\right)
\end{align}

\section{Experiments and Results}

\subsection{Experimental Settings}
\textbf{Dataset.} We used Medical Information Mart for Intensive Care III (MIMIC-III)~\cite{johnson2016mimic} dataset to validate our approach. In order to train a model that is disease-specific, we extracted all admissions that contain at least one diagnostic ICD code related to heart disease. In total 5875 admissions from 1958 patients were used for training. Further, clinical notes were exported from the "NOTEEVENTS" table from the dataset. As for the test set, clinical notes from 25 admissions that not in the training set were examined and labeled by a heart-disease physician with over 15-year experience. 

\textbf{Note Pairing.}  For note pair $D^i$ and $D^{j}$, $i< j \leq n$, we required their time span is at least six months. 

\textbf{Baselines.} Since our approach does not require any external annotations, our baselines are all unsupervised. The first one is \texttt{Most-Entity (ME)}, which greedily picks sentences with most medical entities. The second baseline is \texttt{TF-IDF}, which weights sentences using the sum of words' TF-IDF scores. To avoid duplicate information that could be raised by greedy search, we incorporated Maximal Marginal Relevance (MMR)~\cite{DBLP:conf/sigir/98} with \texttt{TF-IDF}, denoted as \texttt{TF-IDF + MMR (TM)}. MMR subtracts similarity between a candidate and already selected sentences in summary from its weight. To make the comparison fair, we constrained all methods to summarize within the \textbf{same word limit}. 

\textbf{Metrics.} ROUGE-1, ROUGE-2 and ROUGE-L~\cite{DBLP:conf/naacl/LinH03, lin2004rouge} were adopted for evaluation, which measure recall-oriented overlap between automatically generated summary and reference. 

\textbf{Implementation Details.}
CliNER~\cite{boag2015cliner} was used to get the medical entity from clinical notes. For ILP,  $L_1$ and $L_2$ are set as $|D^i|/5.5$ and $|D^i|/4$ respectively. For the neural model, both word embedding and Bi-GRU have 200 dimensions. Batch size is 16. We used Adam as our optimizer with learning rate $10^{-3}$.
\subsection{Results and Discussions} 
Table \ref{tab:results} shows the evaluation results for \texttt{Most-Entity}, \texttt{TF-IDF}, \texttt{TF-IDF + MMR} and \texttt{Ours}. Our method achieved the best performance over all three metrics. We observe that redundancy has high impact on the performance. Without \texttt{MMR} and \texttt{novelty}, both \texttt{TF-IDF} and \texttt{ours} degraded significantly in performance. We also notice that the \texttt{position} term improves the performance of \texttt{ours}. This is within our expectation, since clinical notes are usually written by following some templates.

\begin{table}[t]
\centering
\caption{Experimental results for summarization}
\label{tab:results}
\scalebox{0.9}{
\begin{tabular}{cccc}
\toprule 
Methods                                      & ROUGE-1                   & ROUGE-2                   & ROUGE-L                   \\  \hline
\texttt{Most Entity (ME)}    & 0.41   & 0.26   & 0.40    \\
\texttt{TF-IDF}   & 0.43  & 0.31  & 0.44  \\ 
\multicolumn{1}{c}{\texttt{TF-IDF + MMR (TM)}}           & \multicolumn{1}{c}{0.49} & \multicolumn{1}{c}{0.35} & \multicolumn{1}{c}{0.48} \\
\multicolumn{1}{c}{\texttt{Ours w/o novelty}}           & \multicolumn{1}{c}{0.48} & \multicolumn{1}{c}{0.33} & \multicolumn{1}{c}{0.47} \\
\multicolumn{1}{c}{\texttt{Ours w/o position}}        & \multicolumn{1}{c}{0.50} & \multicolumn{1}{c}{0.36} & \multicolumn{1}{c}{0.49} \\
\multicolumn{1}{c}{\texttt{Ours} } & \multicolumn{1}{c}{\textbf{0.53
}} & \multicolumn{1}{c}{\textbf{0.38}} & \multicolumn{1}{c}{\textbf{0.51}} \\ 
\bottomrule
\end{tabular}
}
\vspace{-1.5em}
\end{table}

\begin{table}[h]
\centering
\caption{Some examples of extracted sentences for summarization by different methods. \texttt{REF} indicates physician's annotation. $\checked$ means the sentence is selected and \texttimes \  means it is not chosen.}
\scalebox{0.80}{
\begin{tabular}{|l|l||llll|}

\hline        
& \textbf{Sentence}  & \texttt{ME} & \texttt{TM} & \texttt{Ours} & \texttt{REF} \\ \hline
 1 & Chronic systolic heart failure   & \texttimes           & \texttimes          & \checked    & \checked         \\ \hline
2 & Clopidogrel 75 mg  & \texttimes           & \texttimes         & \texttimes    & \checked         \\ \hline
3 & \begin{tabular}[c]{@{}l@{}}Facility with copd, lifelong current tobacco abuse\end{tabular}   & \texttimes           & \checked          & \checked    & \checked         \\ \hline

 4 & \begin{tabular}[c]{@{}l@{}}Atrovent hfa 17 mcg/actuation aerosol sig one (1)\end{tabular}    & \texttimes            & \checked           & \texttimes    & \texttimes         \\ \hline
 
5 & \begin{tabular}[c]{@{}l@{}} \small{Please draw vanco level, hct, bun, creatinine on and call results to doctor.} \end{tabular}                                      & \checked    & \checked         & \texttimes     & \texttimes 
\\
\hline

 6 & \begin{tabular}[c]{@{}l@{}}PT was admitted with presumed CHF exacerbation most likely secondary to the increased \\ dietary salt intake and severe aortic stenosis.\end{tabular} & \checked            & \checked           & \checked     & \checked \\ \hline
\end{tabular}
}
\vspace{-0.5em}
\label{tab:examples}
\end{table}

Table \ref{tab:examples} shows examples for sentences predicted by \texttt{ME}, \texttt{TM} and \texttt{Ours}. We have interesting findings.

\texttt{TM} and \texttt{ME} have a strong bias towards long sentences with more entities. They failed to extract important diagnoses such as No.1, which is short in length. No.2 shows that our method also prefers long sentences for the prescription, due to the use of log-sum-exp trick, but not as biased as \texttt{TM} and \texttt{ME}. For No.3, CliNER only extracted one entity "copd", which led $\texttt{ME}$ to make a wrong decision. For No.4, several infrequent terms increased its TF-IDF weight. However, those terms are not very relevant to heart disease and REF actually excludes this sentences. 

Learning from pseudo-labels, our method is capable of determining which entities are more relevant to a disease, heart disease in this case. Both \texttt{TM} and \texttt{ME} falsely selected No.5, while \texttt{Ours} considers it as unimportant correctly. This sentence is actually an instruction for patients themselves and not very important for physicians. No.6 is an example where all methods worked perfectly.

\section{Conclusion}
In this work, we studied the problem of summarizing EHRs and explored three research questions. For \textbf{RQ1}, we proposed to utilize medial entities to cover the intrinsic correlation between multiple EHRs for one patient. For \textbf{RQ2}, we developed an optimization target and used ILP to generate pseudo-labels, which requires no external human supervision. Then for \textbf{RQ3}, those pseudo-labels helped us train a supervised extractive neural model, where the RNN increases the ability to understand contexts and distinguish irrelevant information. We also proposed and validated that adding novelty features to avoid duplicates and considering the position of sentences are significant in summarizing EHRs. Experiments showed that our method outperforms existing unsupervised baselines, which has great potential in helping physicians better understand patients' medical history, reducing costs and improving the quality of patient care.

\section*{Acknowledgement}
The authors thank Hongyang Zhang, Yaodong Yu, Jiacheng Xu, Shikun Zhang and Sean Chen for valuable help.



 
\bibliographystyle{abbrv}
\bibliography{reference}
\end{document}